\documentclass[conference]{IEEEtran}
\IEEEoverridecommandlockouts

\usepackage{cite}
\usepackage{amsmath,amssymb,amsfonts,amsthm}
\usepackage{pifont}
\usepackage{graphicx}
\usepackage{textcomp}
\usepackage{xcolor}
\usepackage{enumitem}

\theoremstyle{plain}
\newtheorem{lemma}{Lemma}
\newtheorem{theorem}{Theorem}

\definecolor{MyGreen}{HTML}{009900}
\definecolor{MyBlue}{HTML}{0d3983}
\RequirePackage[colorlinks,citecolor=MyGreen]{hyperref}
\usepackage{cleveref}

\def\BibTeX{{\rm B\kern-.05em{\sc i\kern-.025em b}\kern-.08em
    T\kern-.1667em\lower.7ex\hbox{E}\kern-.125emX}}
\begin{document}

\title{ERGNN: Spectral Graph Neural Network With Explicitly-Optimized Rational Graph Filters}

\author{\IEEEauthorblockN{Guoming Li}
\IEEEauthorblockA{\textit{Dept. of Machine Learning}\\
\textit{MBZUAI} \\
Abu Dhabi, UAE \\
Paskardli@outlook.com}
\and
\IEEEauthorblockN{Jian Yang}
\IEEEauthorblockA{\textit{Institute of Automation}\\
\textit{Chinese Academy of Sciences}\\
Beijing, China \\
jianyang0227@gmail.com}
\and
\IEEEauthorblockN{Shangsong Liang \thanks{Corresponding authors: Guoming Li and Shangsong Liang.}}
\IEEEauthorblockA{
\textit{MBZUAI, Abu Dhabi, UAE} \\
\textit{Sun Yat-sen University, Guangzhou, China} \\
Guangzhou, China \\
liangshangsong@gmail.com}
}
\maketitle
\begin{abstract}
Approximation-based spectral graph neural networks, which construct graph filters with function approximation, have shown substantial performance in graph learning tasks. 
Despite their great success, existing works primarily employ polynomial approximation to construct the filters, whereas another superior option, namely ration approximation, remains underexplored. 
Although a handful of prior works have attempted to deploy the rational approximation, their implementations often involve intensive computational demands or still resort to polynomial approximations, hindering full potential of the rational graph filters. 
To address the issues, this paper introduces ERGNN, a novel spectral GNN with explicitly-optimized rational filter. 
ERGNN adopts a unique two-step framework that sequentially applies the numerator filter and the denominator filter to the input signals, thus streamlining the model paradigm while enabling explicit optimization of both numerator and denominator of the rational filter. 
Extensive experiments validate the superiority of ERGNN over state-of-the-art methods, establishing it as a practical solution for deploying rational-based GNNs. 
\end{abstract}

\section{Introduction}
\label{section-introduction}
Spectral graph neural networks (GNNs) is a distinct branch of GNNs that treat graph data as signals on graph and manipulate these signals in the spectral domain using crucial graph filters~\cite{surveyspectralgnn}. 
Approximation-based ones, which construct graph filters with function approximation, are routinely adopted in practice for their substantial performances in graph learning tasks~\cite{ARMA-RationalGNN,BernNet-GNN-narrowbandresults-1,JacobiConv,decoupled-PCConv}. 
Prior studies have revealed that GNNs benefit from more accurate filter approximations, prompting the introduction of diverse approximation techniques to build more potent approximation-based GNNs~\cite{GPRGNN,BernNet-GNN-narrowbandresults-1,ChebNetII,chebnet2d,trigonet,JacobiConv,OptBasisGNN,decoupled-PCConv}. 

However, prior research primarily employs {\it polynomial approximation} to construct the graph filters~\cite{GPRGNN,BernNet-GNN-narrowbandresults-1,JacobiConv,ChebNetII,OptBasisGNN,chebnet2d,decoupled-PCConv}. 
In contrast, {\it rational approximation}, which has provable superiority over polynomials in approximating complex functions~\cite{rationalbetter-2,rationalbetter-5}, has not yet received sufficient attention. 
While a few prior works have attempted to deploy rational approximation, their implementations either involve intensive computation costs~\cite{CayleyNet-RationalGNN} or still resort to polynomials~\cite{ARMA-RationalGNN}, hindering practical application and full potential of the rational graph filters. 

To address these concerns, we propose ERGNN ({\bf E}xplicitly-optimized {\bf R}ational {\bf G}raph {\bf N}eural {\bf N}etwork). 
ERGNN employs a unique two-step framework that sequentially applies the numerator filter and the denominator filter to the input signals. 
Such framework ensures an efficient and effective implementation of the rational graph filter, while also enabling explicit optimization of the filter's numerator and denominator, offering notable flexibility in filter construction. 
In summary, our contributions can be listed below: 

\begin{itemize}[parsep=2pt,itemsep=2pt,topsep=2pt]
\item We propose ERGNN with a streamlined framework that avoids intricate computations required by prior methods and enables explicit optimization of the rational graph filter, offering a practical solution for deploying the rational-based GNNs. 
\item We provide an in-depth analysis of ERGNN, guaranteeing the efficacy of the rational graph filter implemented through the novel two-step framework. 
\item We conduct extensive experiments to validate the superiority of our ERGNN over state-of-the-art methods. The compelling results confirm the accuracy of our analysis and lay the groundwork for future research. 
\end{itemize}

\nocite{fang2021structure}



\begin{figure*}[!t]
\centering
\includegraphics[width=.85\linewidth]{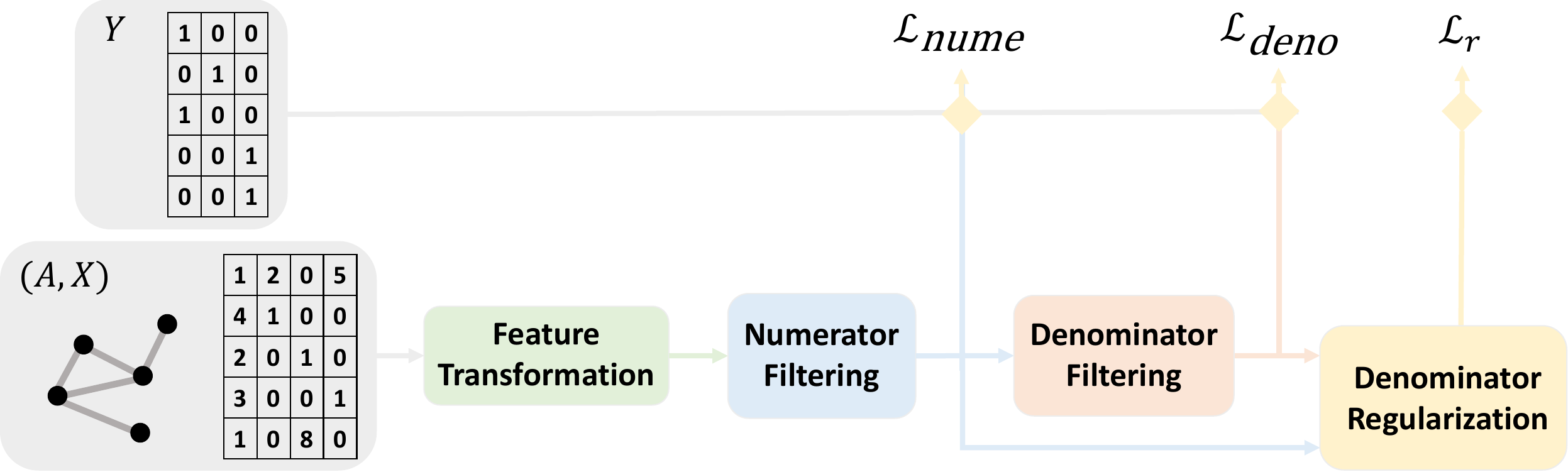}
\vskip -0.05in
\caption{Overall framework of ERGNN.}
\label{fig:ERGNN}
\end{figure*}

\section{Background and Preliminaries}
\label{section-preliminaries}
Let $\mathcal{G}$ be an undirected unweighted graph with adjacency matrix $\boldsymbol{A}$ and normalized graph Laplacian $\boldsymbol{L}$. 
We denote $\boldsymbol{L}=\boldsymbol{U}diag(\boldsymbol{\lambda})\boldsymbol{U}^{T}$ be eigendecomposition of $\boldsymbol{L}$, with $\boldsymbol{U}$, $\boldsymbol{\lambda}$ being the eigenvectors and eigenvalues of $\boldsymbol{L}$, respectively. 

\subsection{Spectral GNNs}
\label{section-preliminaries-sepctral-gnns}
The core of spectral GNNs is to perform filtering on graph signals by leveraging pivotal graph filters. 
To be specific, given a graph signal $\boldsymbol{x}\in\mathbb{R}^{N}$ on $\mathcal{G}$, the filtering operation is given by $f \star \boldsymbol{x} = \boldsymbol{U}diag(f(\boldsymbol{\lambda}))\boldsymbol{U}^{T}\boldsymbol{x}$, where $f(\cdot)$ is the graph filter and $\star$ acts as filtering. 
In practical scenarios, the graph signals operated by spectral GNNs are mostly of the matrix-type. 
Thus, we let $\boldsymbol{X}\in\mathbb{R}^{N\times d}$ be the graph signals, the spectral GNNs are formulated as follows: 
\begin{align}
\label{eq-spectralgnns}
\boldsymbol{Z}=\boldsymbol{U}diag(f(\boldsymbol{\lambda}))\boldsymbol{U}^{T}\boldsymbol{X}\ ,
\end{align}
where $\boldsymbol{Z}$ denotes the output of spectral GNNs. 
In this paper, $\boldsymbol{Z}$ is considered as label prediction of nodes. 

{\bf Approximation-based approaches.} In practice, spectral GNNs are predominantly built upon function approximation techniques. 
These approximation-based approaches typically formulate $f(\cdot)$ with polynomial approximations~\cite{polyapprox_1,polyapprox_2,polyapprox_application} or rational approximation~\cite{rationalbetter-1,rationalbetter-2,rationalbetter-3,rationalbetter-4,rationalbetter-5}, resulting in Eq.~\ref{eq-spectralgnns} being derived into the following forms: 
\begin{align}
\label{eq-spectralgnnapprox-1}
&\text{Polynomial-based :}\quad \boldsymbol{Z}=\sum_{k=0}^{K}\gamma_{k}\boldsymbol{L}^{k}\boldsymbol{X}\ ,\\
\label{eq-spectralgnnapprox-2}
&\text{Rational-based :}\quad \boldsymbol{Z}=\frac{\sum_{k_{1}=0}^{K_{1}}\alpha_{k_{1}}\boldsymbol{L}^{k_{1}}}{\sum_{k_{2}=0}^{K_{2}}\beta_{k_{2}}\boldsymbol{L}^{k_{2}}}\boldsymbol{X}\ .
\end{align}
Here, $\alpha$, $\beta$ and $\gamma$ are designated as learnable parameters, and $K_1$ is generally equal to $K_2$. 
As shown with the equations, in contrast to polynomial-based methods that involve only matrix addition and multiplication, rational-based methods require further computation-intensive matrix inversion. 
This makes existing proposals either involve intensive computations~\cite{CayleyNet-RationalGNN} or resort to polynomial approximations~\cite{ARMA-RationalGNN}, impeding the substantial potential of rational graph filters. 

\subsection{Augmented MLPs as Substitutions of GNNs}
\label{section-preliminaries-gnn-to-mlp}
Recent attempts to integrate graph information into Multi-layer Perceptrons (MLPs) have led to the creation of {\it graph-augmented MLPs}. 
These methods use only augmented node features as input and employ a pure MLP model to generate output embeddings, yet they achieve performance comparable to sophisticated GNNs. 
Existing proposals are categorized into regularization-based and distillation-based methods. {\bf Regularization-based methods} train an MLP encoder via combining supervised loss with a regularization that accounts for the graph structure~\cite{MLP-equal-GNN-reg-2,MLP-equal-GNN-reg-9}; {\bf Distillation-based methods} leverage KL-divergence loss to distill the predictions of a GNN teacher into an MLP student, with the objective of aligning the MLP's predictions with those of the GNN teacher~\cite{MLP-equal-GNN-3,MLP-equal-GNN-4}.

\section{Methodology}
\label{section-method}

\subsection{Model Architecture of ERGNN}
\label{section-method-ergnn-architecture}
Initially, let $\boldsymbol{X}$ be the raw node feature of $\mathcal{G}$, drawing from prior proposals on linear spectral GNNs~\cite{JacobiConv,decoupled-PCConv}, we first employ a linear transformation on $\boldsymbol{X}$: 
\begin{align}
\label{eq-pre-mlp-feature-trans}
\boldsymbol{Z}^{(0)}=\boldsymbol{X}\boldsymbol{W} + \boldsymbol{b}\ .
\end{align}
This $\boldsymbol{Z}^{(0)}$ will serve as the input signals of the unique two-step framework. 
Next, as indicated in Eq.~\ref{eq-spectralgnnapprox-2}, the filtering operation of the rational graph filter is equivalent to the sequential application of numerator filter and denominator filter. 
This motivates our design of ERGNN as a two-step filtering: 

{\bf Step\ding{182}: Numerator filtering.} Evidently, the numerator filtering can be achieved with a polynomial-based graph filter. 
We further generalize the $\sum_{k_{1}=0}^{K_{1}}\alpha_{k_{1}}\boldsymbol{L}^{k_{1}}$ to a unified polynomial approximation paradigm, leading to the following formulation for numerator filtering: 
\begin{align}
\label{eq-numerator-filtering}
\boldsymbol{Z}^{(1)}=\sum_{k_{1}=0}^{K_{1}}\alpha_{k_{1}}\mathrm{T}_{k_{1}}\left(\boldsymbol{L}\right)\boldsymbol{Z}^{(0)}\ ,
\end{align}
where $\mathrm{T}_{k_{1}}\left(\cdot\right)$ denotes the $k_{1}$-order polynomial basis. 

{\bf {\bf Step}\ding{183}: MLP-based denominator filtering.} To circumvent the matrix inversion operation inherent in denominator filtering, inspired by graph-augmented MLPs, we opt for an MLP to substitute the denominator graph filter, as shown below:  
\begin{align}
\label{eq-denominator-filtering}
\boldsymbol{Z}^{(2)}=\text{MLP}_{\mathcal{G}}(\boldsymbol{Z}^{(1)})\ .
\end{align}
The $\boldsymbol{Z}^{(2)}$ will serve as the final output of the ERGNN. 
While $\text{MLP}_{\mathcal{G}}(\cdot)$ does not reflect the form of the denominator filter, a unique {\it regularization} will be introduced into the model optimization to shape this MLP into the structure of a polynomial filter. 
More discussions are detailed in Section~\ref{section-method-loss}. 

\begin{table*}[!t]
  \caption{Experimental results on homophilic and heterophilic datasets: Mean $(\%)$ $\pm$ standard deviation. 
  $\Delta_{\uparrow}$ denotes the improvement of ERGNN relative to the runner-up.} 
  \vskip -0.05in
  \label{table-result-node-classify}
  \centering
  \setlength{\tabcolsep}{1pt}
  \renewcommand\arraystretch{1.05}
  \resizebox{\textwidth}{!}{
  \begin{tabular}{lccccc|ccccc}
    \hline
    Method   &  Cora  &  Cite.  &  Pubm.  &  Comp.  &  Photo &  Actor  &  -empire  &  Tolo. & -ratings & Ques. \\ \hline
    GCN~\cite{GCN}      &  $87.11_{\pm0.75}$  &  $77.56_{\pm0.69}$  &  $87.05_{\pm0.62}$  &  $86.25_{\pm0.77}$  &  $88.54_{\pm0.55}$ & $35.12_{\pm1.61}$ & $55.78_{\pm0.43}$ & $76.65_{\pm0.87}$ & $47.38_{\pm0.47}$ & $71.55_{\pm0.71}$ \\
    GCNII~\cite{gcnii}      &  $88.55_{\pm0.62}$  &  $76.83_{\pm0.71}$  &  $87.86_{\pm0.77}$  &  $89.13_{\pm0.53}$  &  $90.68_{\pm0.47}$  & $37.81_{\pm0.82}$ & $66.25_{\pm0.56}$ & $75.41_{\pm0.44}$ & $43.62_{\pm0.52}$ & $70.71_{\pm0.83}$ \\
    H2GCN~\cite{H2GCN} &  $88.24_{\pm0.42}$  &  $75.93_{\pm0.89}$  &  $88.06_{\pm0.48}$  &  $89.55_{\pm0.57}$  &  $90.13_{\pm0.52}$  & $38.54_{\pm0.75}$ & $70.73_{\pm0.81}$ & $77.62_{\pm0.59}$ & $46.52_{\pm0.70}$ & $71.19_{\pm0.57}$\\
    APPNP~\cite{decoupled-advantages-3-coupled-disadvantages-2-APPNP}    &  $87.79_{\pm1.32}$  &  $77.62_{\pm1.63}$  &  $88.46_{\pm0.82}$  &  $87.11_{\pm0.93}$  &  $89.43_{\pm0.63}$ & $37.66_{\pm1.04}$  & $72.74_{\pm0.63}$ & $73.92_{\pm0.60}$ & $48.41_{\pm0.34}$  & $72.08_{\pm0.91}$  \\
    Nodeformer~\cite{nodeformer}      &  $88.21_{\pm0.88}$  &  $76.72_{\pm0.85}$  &  $88.10_{\pm0.67}$  &  $89.28_{\pm0.63}$  &  $90.71_{\pm0.48}$ & $38.81_{\pm0.73}$  & $73.42_{\pm0.63}$ & $79.47_{\pm0.48}$ & $45.48_{\pm0.50}$ & $68.51_{\pm0.96}$ \\
    GLOGNN~\cite{glognn++}      &  $88.76_{\pm0.62}$  &  $78.18_{\pm0.52}$  &  $88.68_{\pm0.44}$  &  $\underline{90.63_{\pm0.28}}$  &  $94.19_{\pm0.71}$ & $39.29_{\pm1.26}$ & $67.39_{\pm0.34}$ & $79.08_{\pm0.72}$ & $\underline{49.88_{\pm0.44}}$  & $74.12_{\pm0.87}$  \\ 
    \hline
    ChebNet~\cite{ChebNet}  &  $86.79_{\pm1.55}$  &  $76.78_{\pm1.92}$  &  $85.43_{\pm0.63}$  &  $89.77_{\pm1.22}$  &  $91.55_{\pm1.23}$ & $33.88_{\pm1.21}$  & $63.69_{\pm0.33}$ & $71.22_{\pm0.53}$ & $38.19_{\pm0.41}$  & $65.13_{\pm1.25}$ \\
    GPRGNN~\cite{GPRGNN}  &  $88.75_{\pm1.07}$  &  $78.15_{\pm1.78}$  &  $88.76_{\pm0.82}$  &  $90.05_{\pm0.68}$  &  $93.49_{\pm0.59}$ & $40.16_{\pm0.87}$  & $73.44_{\pm0.35}$ & $75.12_{\pm0.64}$ & $49.56_{\pm0.47}$    & $73.82_{\pm1.08}$  \\
    BernNet~\cite{BernNet-GNN-narrowbandresults-1}  &  $88.49_{\pm1.22}$  &  $78.03_{\pm1.32}$  &  $\underline{89.47_{\pm0.47}}$  &  $90.12_{\pm0.72}$  &  $92.51_{\pm0.71}$ & $40.57_{\pm0.93}$ & $72.66_{\pm0.81}$ & $74.71_{\pm0.63}$ & $48.73_{\pm0.43}$    & $73.65_{\pm1.14}$  \\
    CayleyNet~\cite{CayleyNet-RationalGNN}  & $88.13_{\pm1.21}$  &  $77.25_{\pm1.78}$  &  $87.81_{\pm0.91}$  &  $88.58_{\pm0.60}$  &  $91.69_{\pm0.68}$ & $34.38_{\pm1.25}$ & $63.18_{\pm0.45}$ & $72.53_{\pm1.02}$ & $41.25_{\pm0.88}$  & $63.12_{\pm0.53}$  \\
    ARMA~\cite{ARMA-RationalGNN} & $88.69_{\pm1.03}$  &  $78.07_{\pm1.53}$  &  $88.30_{\pm0.81}$  &  $87.81_{\pm0.54}$  &  $93.13_{\pm0.47}$ & $38.11_{\pm1.31}$ & $68.28_{\pm0.73}$ & $73.82_{\pm0.73}$ & $42.09_{\pm0.62}$  & $69.24_{\pm0.48}$ \\
    JacobiConv~\cite{JacobiConv} & $89.19_{\pm1.08}$  &  $78.33_{\pm1.15}$  &  $89.19_{\pm0.58}$  &  $90.19_{\pm0.59}$  &  $\underline{94.88_{\pm0.32}}$ & $40.64_{\pm1.17}$ & $74.35_{\pm0.65}$ & $78.41_{\pm0.38}$ & $48.56_{\pm0.22}$   & $73.78_{\pm0.75}$  \\
    ChebNetII~\cite{ChebNetII}  & $88.72_{\pm0.76}$  &  $\underline{78.47_{\pm1.12}}$  &  $88.75_{\pm0.89}$  &  $90.15_{\pm0.73}$  &  $94.30_{\pm0.54}$ & $41.03_{\pm1.28}$ & $\underline{74.58_{\pm0.35}}$ & $79.02_{\pm0.60}$ & $49.43_{\pm0.41}$  & $73.93_{\pm0.69}$   \\
    OptBasis~\cite{OptBasisGNN}  & $88.76_{\pm1.03}$  &  $78.38_{\pm1.18}$  &  $89.35_{\pm0.63}$  &  $90.27_{\pm1.03}$  &  $94.71_{\pm0.33}$ & $\underline{41.11_{\pm0.75}}$ & $74.29_{\pm0.33}$ & $79.25_{\pm0.53}$ & $49.48_{\pm0.36}$  & $\underline{74.16_{\pm0.75}}$ \\
    PCNet~\cite{decoupled-PCConv} &  $\underline{89.23_{\pm0.82}}$  &  $78.22_{\pm0.91}$  &  $87.72_{\pm0.55}$  &  $90.02_{\pm0.67}$  &  $94.54_{\pm0.82}$ & $40.39_{\pm0.73}$ & $73.17_{\pm0.51}$ & $77.63_{\pm0.41}$ & $48.77_{\pm0.57}$ & $73.97_{\pm0.70}$ \\ 
    \hline
    ERGNN({\bf ours}) & $\boldsymbol{90.41_{\pm1.16}}$  &  $\boldsymbol{79.25_{\pm0.73}}$  &  $\boldsymbol{91.11_{\pm0.65}}$  &  $\boldsymbol{93.12_{\pm0.55}}$  &  $\boldsymbol{96.87_{\pm0.48}}$ & $\boldsymbol{43.05_{\pm0.62}}$ & $\boldsymbol{77.67_{\pm0.81}}$ & $\boldsymbol{82.88_{\pm0.58}}$ & $\boldsymbol{53.69_{\pm0.54}}$ & $\boldsymbol{78.39_{\pm0.70}}$ \\
    $\Delta_{\uparrow}$ & $1.18\%$ & $0.78\%$ & $1.64\%$ & $2.49\%$ & $1.99\%$ & $1.94\%$ & $3.09\%$ & $3.22\%$ & $3.81\%$ & $4.23\%$\\
    \hline
  \end{tabular}}
\end{table*}

\subsection{Optimization Target}
\label{section-method-loss}
Suppose $\boldsymbol{Y}$ be the true node label. 
To ensure the proper construction of the rational graph filter in ERGNN and its performance on downstream tasks, we propose to optimize an innovative loss function $\mathcal{L}$ defined as below:
\begin{align}
\label{eq-loss-function}
&\mathcal{L}=\eta\mathcal{L}_{\text{nume}} + \xi\mathcal{L}_{\text{deno}} + \mathcal{L}_{\text{r}}\ \\
s.t.\quad &\mathcal{L}_{\text{nume}}=\text{CE}\left(\boldsymbol{Z}^{(1)}, \boldsymbol{Y}\right)\ \\
&\mathcal{L}_{\text{deno}}=\text{CE}\left(\boldsymbol{Z}^{(2)}, \boldsymbol{Y}\right)\ \\
&\mathcal{L}_{\text{r}}=\text{CE}\left(\boldsymbol{Z}^{(1)}, \sum_{k_{2}=0}^{K_{2}}\beta_{k_{2}}\mathrm{T}_{k_{2}}\left(\boldsymbol{L}\right)\boldsymbol{Z}^{(2)}\right)\ ,
\end{align}
where $\eta$ and $\xi$ are trade-off parameters, and $\text{CE}(\cdot,\cdot)$ denotes cross-entropy loss. 
Both $\mathcal{L}_{\text{nume}}$ and $\mathcal{L}_{\text{deno}}$ measure the prediction error of the outputs from the two filtering operations. 
The regularization term $\mathcal{L}_{\text{r}}$ ensures the equivalence of $\text{MLP}_{\mathcal{G}}(\boldsymbol{Z}^{(1)})$ with a polynomial denominator filtering. 
The overall framework of ERGNN is depicted in Figure~\ref{fig:ERGNN}. 
In the following section, we will provide in-depth theoretical explanation on the rationale and validity of these loss terms. 

\section{Theoretical Insights}
\label{section-theory}
We now provide in-depth theoretical analysis to substantiate the soundness and effectiveness of our ERGNN. 
To begin with, as noted in previous studies~\cite{GPRGNN,BernNet-GNN-narrowbandresults-1,JacobiConv,ChebNetII,OptBasisGNN,decoupled-PCConv}, the objective of approximation-based spectral GNNs is to use the approximate construction $f(\boldsymbol{\lambda})$ to approximate the target (optimal) filter $f^{*}(\boldsymbol{\lambda})$, with better approximations leading to superior model performance of spectral GNNs. 
Building on this insight, we examine the approximation capability of the rational graph filter in ERGNN compared to polynomial graph filters. 
Specially, we introduce Appel's method on best-fit rational approximation~\cite{fixed_numerator_learned_denominator-1,fixed_numerator_learned_denominator-3}, as follows: 
\begin{lemma}
\label{lemma-appel}
(\textbf{Restatement of Appel's method}~\cite{fixed_numerator_learned_denominator-1}) Let decay-type function $F(x)$ be the approximation target. 
Given a fixed function $P(x)$, we can derive a polynomial function $Q(x)=\sum_{i=1}^{n}w_{i}x^{i}$ with coefficients $w_{i}$ computed via Appel's algorithm, resulting in $\frac{P(x)}{Q(x)}$ as a rational approximation of $F(x)$ that satisfies the Chebyshev best-fit~\cite{polyapprox_1}. 
\end{lemma}
Lemma~\ref{lemma-appel} indicates that even with a fixed numerator function, we can obtain the Chebyshev best-fit rational approximation by adjusting only the denominator polynomial. 
This insight leads us to propose the following theorem: 
\begin{theorem}
\label{theorem-appel}
Let $f^{*}(\boldsymbol{\lambda})$ be the target (optimal) filter we aim to approximate, and let $\sum_{k_{1}=0}^{K_{1}}\alpha^{*}_{k_{1}}\mathrm{T}_{k_{1}}\left(\boldsymbol{\lambda}\right)$ be the optimal approximation to $f^{*}(\boldsymbol{\lambda})$ using polynomial $\mathrm{T}$. 
By considering a rational graph filter of the form $\frac{\sum_{k_{1}=0}^{K_{1}}\alpha^{*}_{k_{1}}\mathrm{T}_{k_{1}}\left(\boldsymbol{\lambda}\right)}{\sum_{k_{2}=0}^{K_{2}}\beta_{k_{2}}\mathrm{T}_{k_{2}}\left(\boldsymbol{\lambda}\right)}$, we can achieve a rational graph filter that satisfies the Chebyshev best-fit by adjusting (learning) the parameters $\beta$. 
\end{theorem}
The theorem is evidently valid. 
To be specific, since $\boldsymbol{\lambda}$ lies within $\left[0,2\right]$, the approximation $f(\boldsymbol{\lambda})$ is equivalent to a decay-type function that approaches $0$ at infinity, thus making Lemma~\ref{lemma-appel} applicable to the graph filters. 

The theorem above offers a theoretical guarantee for the two-step framework and optimization strategy of ERGNN. 
In detail, ERGNN learns the optimal numerator filter via the loss $\mathcal{L}_{\text{nume}}$ and also optimizes the denominator filter with the fixed numerator using $\mathcal{L}_{\text{deno}}$ and $\mathcal{L}_{\text{r}}$, resulting in the rational graph filter that satisfies the Chebyshev best-fit. 
Furthermore, since the obtained numerator filter is the optimal polynomial approximation to the target $f^{*}(\boldsymbol{\lambda})$, the rational filter provably offers a better approximation to $f^{*}(\boldsymbol{\lambda})$ than polynomial ones.

\begin{table*}[!t]
  \caption{Experimental results on large graphs: Mean $(\%)$ $\pm$ standard deviation. 
  OOM denotes ``out of memery''. 
  Complexity on papers100M consists of: average training time per epoch (second) / total training time (hours) / number of parameters (kilo).} 
  \vskip -0.05in
  \label{table-result-large}
  \centering
  \setlength{\tabcolsep}{6.5pt}
  \renewcommand\arraystretch{1.05}
  \resizebox{\textwidth}{!}{
  \begin{tabular}{lccc|ccc|c}
    \hline  
    Method & arxiv  &  products & papers100M &  Penn94  &  Genius  &  Gamers & Complexity on papers100M \\ \hline  
    GCN~\cite{GCN}  & $71.74_{\pm0.33}$ & $75.64_{\pm0.27}$ & OOM & $82.47_{\pm0.36}$ & $87.42_{\pm0.34}$ & $62.18_{\pm0.32}$ & - \\
    Nodeformer~\cite{nodeformer} & $65.08_{\pm0.61}$ & $73.66_{\pm0.52}$ & OOM & $83.71_{\pm0.34}$ & $89.38_{\pm0.48}$ & $63.18_{\pm0.58}$ & - \\
    LINKX~\cite{dataset6-large-hetero} & $67.23_{\pm0.34}$ & $77.52_{\pm0.70}$ & OOM & $85.15_{\pm0.17}$ & $90.69_{\pm0.44}$ & $66.48_{\pm0.32}$ & - \\
    GloGNN~\cite{glognn++}  & $72.11_{\pm0.54}$ & $81.51_{\pm0.43}$ & $66.73_{\pm0.41}$ &  $\underline{85.54_{\pm0.31}}$ & $90.72_{\pm0.26}$ & $\underline{66.52_{\pm0.42}}$  & $17.8$ / $2.73$ / $4,495$K \\ 
    \hline
    GPRGNN~\cite{GPRGNN}  & $71.96_{\pm0.47}$ & $80.75_{\pm0.35}$ & $65.66_{\pm0.38}$ &  $83.87_{\pm0.65}$ & $90.07_{\pm0.43}$ & $63.18_{\pm0.61}$ & $15.1$ / $2.70$ / $4,495$K \\ 
    BernNet~\cite{BernNet-GNN-narrowbandresults-1}  & $71.73_{\pm0.24}$ & $79.67_{\pm0.36}$ & $65.12_{\pm0.32}$ &  $84.18_{\pm0.52}$ & $89.69_{\pm0.52}$ & $62.93_{\pm0.38}$ & $15.3$ / $2.54$ / $4,495$K \\ 
    JacobiConv~\cite{JacobiConv}  & $72.18_{\pm0.21}$ & $81.75_{\pm0.42}$ & $64.28_{\pm0.24}$ &  $84.56_{\pm0.42}$ & $90.63_{\pm0.35}$ & $65.23_{\pm0.37}$ & $9.3$ / $1.78$ / $4,700$ \\ 
    ChebNetII~\cite{ChebNetII}  & $\underline{72.33_{\pm0.34}}$ & $81.07_{\pm0.36}$ & $\underline{67.21_{\pm0.33}}$ &  $85.22_{\pm0.36}$ & $90.47_{\pm0.28}$ & $65.71_{\pm0.24}$ & $15.5$ / $2.38$ / $4,495$K \\ 
    OptBasis~\cite{OptBasisGNN}  & $72.26_{\pm0.42}$ & $81.15_{\pm0.33}$ & $66.82_{\pm0.36}$ &  $84.82_{\pm0.60}$ & $\underline{90.91_{\pm0.24}}$ & $65.86_{\pm0.44}$  & $16.1$ / $2.62$ / $4,495$K \\ 
    PCNet~\cite{decoupled-PCConv} & $72.08_{\pm0.21}$ & $\underline{82.15_{\pm0.32}}$ & $65.63_{\pm0.27}$ &  $83.75_{\pm0.40}$ & $90.33_{\pm0.45}$ & $65.28_{\pm0.58}$  & $15.8$ / $2.64$ / $4,495$K \\
    \hline
    ERGNN({\bf ours}) & $\boldsymbol{73.29_{\pm0.38}}$ & $\boldsymbol{83.37_{\pm0.42}}$ & $\boldsymbol{69.08_{\pm0.47}}$ &  $\boldsymbol{86.91_{\pm0.50}}$ & $\boldsymbol{92.25_{\pm0.32}}$ & $\boldsymbol{67.98_{\pm0.39}}$  & $\boldsymbol{11.8}$ / $\boldsymbol{2.35}$ / $\boldsymbol{4,495}$K \\
    $\Delta_{\uparrow}$ & $0.96\%$ & $1.22\%$ & $1.87\%$ & $1.37\%$ & $1.34\%$ & $1.46\%$ & - \\
    \hline
  \end{tabular}}
\end{table*}

\section{Empirical Studies}
\label{section-experiments}
We now perform extensive experiments to showcase the superior efficacy of ERGNN against state-of-the-art baselines. 

\subsection{Node Classification on Real-world Graphs}
\label{section-experiments-node-classification}
{\bf Experimental settings.} We perform routine node classification experiments on $10$ benchmark datasets, including $5$ {\it homophilic} datasets: Cora, CiteSeer and PubMed~\cite{dataset1-cora}, Computers and Photo~\cite{dataset2-photo-comp}; and $5$ {\it heterophilic} datasets: Actor~\cite{dataset7-actor}, Roman-empire, Tolokers, Amazon-ratings and Questions~\cite{dataset8-small-hetero}. 
The selected datasets deviate from the routine choices that have been flagged for data-leakage issues, as proven by~\cite{dataset8-small-hetero}. 
We adopt the routine $60\%/20\%/20\%$ split ratio for Actor and homophilic datasets following~\cite{BernNet-GNN-narrowbandresults-1,ChebNetII,JacobiConv,OptBasisGNN,decoupled-PCConv}, and use $50\%/25\%/25\%$ ratio for the remaining datasets following~\cite{dataset8-small-hetero}. 
We evaluate each model on $10$ pre-generated splits of each dataset and use Adam optimizer~\cite{Adamoptimizer} for model training, with early stopping $250$ and maximum epochs $2000$. 

Baseline models include $9$ spectral GNNs and $6$ non-spectral methods, ensuring extensive coverage of baselines. 
For our ERGNN, we use Chebyshev interpolation~\cite{chebyshevinterpolation} and set both $K_1$ and $K_2$ to $10$ following ChebNetII~\cite{ChebNetII}. 
We set $\text{MLP}_{\mathcal{G}}$ to be a 2-layer MLP with $64$ hidden units. 
Learning rate, weight decay and dropout are set within same search range as other baselines~\cite{BernNet-GNN-narrowbandresults-1,ChebNetII,JacobiConv,OptBasisGNN,decoupled-PCConv}. 

{\bf Results and analysis.} We report the results in Table~\ref{table-result-node-classify}, with the best performance highlighted in \textbf{bold}, and the runner-up \underline{underlined}. 
We observe that our ERGNN consistently outperforms all baselines across both homophilic and heterophilic datasets, with performance gains over the runner-up methods reaching up to $4.23\%$. 
This not only underscores the superiority of our proposed approach but also serves as empirical validation of the correctness of our analysis.

\subsection{Evaluating Scalability on Large Graphs}
\label{section-experiments-node-classification-large-graphs}
{\bf Experimental settings.} We assess the scalability of our ERGNN through experiments on large graphs, including $3$ homophilic graphs: papers100M, products and arxiv~\cite{dataset5-ogb}; and $3$ heterophilic graphs: Penn94, Genius and Gamers~\cite{dataset6-large-hetero}. 
We generate $5$ random splits for each dataset using the partitions suggested in~\cite{dataset5-ogb,dataset6-large-hetero} and apply batch training. 

For ERGNN, we search $K_{1}$ ($=K_{2}$) over $\{4,6,8,10,12\}$, and optimize $\text{MLP}_{\mathcal{G}}$ with number of layers on $\{2,3\}$ and hidden units among $\{128,256,512,1024,2048\}$. 

{\bf Results and analysis.} As outlined in Table~\ref{table-result-large}, our ERGNN outperforms all baselines across various large datasets, showcasing notable enhancements compared to the runner-up results (up to a remarkable $1.87\%$ increase). 
Moreover, as evidenced by the model complexity evaluation on papers100M, ERGNN surpasses most baselines in both computational and parameter complexity. 
These results underscore the exceptional prowess of ERGNN in effectively handling large-scale datasets characterized by intricate patterns, validating the outstanding scalability of our proposal. 

\subsection{Learning Graph Filters from Signals}
\label{section-experiments-learning-graph-filters}
{\bf Experimental settings.} Following previous studies~\cite{BernNet-GNN-narrowbandresults-1,JacobiConv,OptBasisGNN}, we assess the filter approximation ability of ERGNN through experiments of learning $5$ graph filters (low: $e^{-10\lambda^2}$, high: $1$$-$$e^{-10\lambda^2}$, band: $e^{-10(\lambda-1)^2}$, reject: $1$$-$$e^{-10(\lambda-1)^2}$ and comb: $\vert\sin(\pi\lambda)\vert$) from processed signals on a grid graph. 
We adopt the same settings for ERGNN as those employed in prior works, with the average squared error of both ERGNN and baselines reported as the evaluation metric. 

{\bf Results and analysis.} As reported in Table~\ref{table-result-numerical-exp}, ERGNN achieves far lower error compared to the counterparts across all $5$ filters (even reducing the error by $94\%$ compared to the runner-up on the reject filter). 
These results not only highlight ERGNN's prowess in filter approximation but also affirm the superiority of rational over polynomial (note that ARMA is actually implemented using polynomial approximations~\cite{ARMA-RationalGNN}). 

\begin{table}[!t]
  \caption{Average squared error across $5$ graph filters: lower indicates better performance.} 
  \vskip -0.05in
  \label{table-result-numerical-exp}
  \centering
  \setlength{\tabcolsep}{5.5pt}
  \renewcommand\arraystretch{1.05}
  \resizebox{\linewidth}{!}{
  \begin{tabular}{lccccc}
    \hline  
     & Low  &  High & Band &  Reject  &  Comb \\ \hline  
    ARMA & $1.8478$ & $1.8632$ & $7.6922$ & $8.2732$ & $15.1214$ \\
    GPRGNN & $0.4169$ & $0.0943$ & $3.5121$ & $3.7917$ & $4.6549$ \\
    BernNet & $0.0314$ & $0.0113$ & $0.0411$ & $0.9313$ & $0.9982$ \\
    PCNet & $0.0076$ & $0.0054$ & $0.0291$ & $0.0371$ & $0.4426$ \\
    ChebNetII & $0.0008$ & $0.0015$ & $0.0255$ & $0.0173$ & $0.3761$ \\
    JacobiConv & $0.0003$ & $0.0011$ & $0.0213$ & $0.0156$ & $0.2933$ \\
    OptBasis & $0.0002$ & $0.0009$ & $0.0227$ & $0.0148$ & $0.2976$ \\
    \hline
    {\bf ERGNN} & $\boldsymbol{0.0000}$ & $\boldsymbol{0.0004}$ & $\boldsymbol{0.0027}$ & $\boldsymbol{0.0009}$ & $\boldsymbol{0.0096}$ \\
    \hline
  \end{tabular}}
\end{table}

\section{Conclusions}
\label{section-conclusion}
This paper proposes ERGNN, an innovative spectral GNN with an explicitly-optimized rational graph filters. 
ERGNN applies a unique two-step framework, sequentially building the numerator filter and the MLP-based denominator filter, thus avoiding intricate computations required by prior methods and enabling an explicitly-optimized rational filter. 
Backed by in-depth theoretical insight, the superiority of our ERGNN are justified. 
Extensive experiments further validate that ERGNN outperforms state-of-the-art methods, highlighting the exceptional efficacy and scalability of ERGNN. 

Our ERGNN with two-step framework offers a viable solution for deploying rational-based GNNs in practical applications. 
Future research can build upon this by proposing more concrete frameworks that allow for direct control \& optimization over the filter constructions. 




\bibliographystyle{IEEEtran}
\bibliography{ieeefull-refs}

\end{document}